\documentclass[runningheads]{llncs}
\usepackage{graphicx}

\usepackage{tikz}
\usepackage{comment}
\usepackage{amsmath,amssymb} 
\usepackage{color}
\usepackage{ltablex}
\usepackage{stmaryrd}
\usepackage{mathtools}
\usepackage{bm}
\usepackage{makecell}
\usepackage{multirow}
\usepackage{hyperref}
\usepackage[titletoc]{appendix}

\newcommand{\matr}[1]{\bm{#1}}     
\newcommand{\zht}[1]{{{#1}}}
\newcommand{\haofu}[1]{{{#1}}}

\newcommand{\RNum}[1]{\uppercase\expandafter{\romannumeral #1\relax}}
\newcommand{\etal}{\textit{et al.}}

\newcommand{\best}[1]{{\bf{#1}}}

\DeclareMathOperator{\G}{G}
\DeclareMathOperator{\D}{D}

\DeclareMathOperator{\softmax}{softmax}

\DeclareMathOperator{\relu}{LReLU}
\DeclareMathOperator{\mlp}{mlp}
\DeclareMathOperator{\gap}{gap}

\begin{document}
\pagestyle{headings}
\def\ECCVSubNumber{2168}  

\title{Example-Guided Image Synthesis across Arbitrary Scenes using Masked Spatial-Channel Attention and Self-Supervision}

\titlerunning{Example-Guided Image Synthesis across Arbitrary Scenes}
%
\author{Haitian Zheng
\quad Haofu Liao \quad Lele Chen \quad Wei Xiong \quad Tianlang Chen \quad Jiebo Luo\\
University of Rochester\\
{\tt\small \{hzheng15, hliao6, lchen63, wxiong5, tchen45, jluo\}@cs.rochester.edu}
}
\author{Haitian Zheng
\quad Haofu Liao \quad Lele Chen \\ Wei Xiong \quad Tianlang Chen \quad Jiebo Luo}
\authorrunning{H. Zheng et al.}
%
\institute{University of Rochester, \\
\email{\tt\small \{hzheng15, hliao6, lchen63, wxiong5, tchen45, jluo\}@cs.rochester.edu}\\
}

\maketitle
\vspace{-4mm}
\begin{abstract}
Example-guided image synthesis has recently been attempted to synthesize an image from a semantic label map and an exemplary image. In the task, the additional exemplar image provides the style guidance that controls the appearance of the synthesized output. Despite the controllability advantage, the existing models are designed on datasets with specific and roughly aligned objects. In this paper, we tackle a more challenging and general task, where the exemplar is an arbitrary scene image that is semantically different from the given label map. 
To this end, we first propose a Masked Spatial-Channel Attention (MSCA) module which models the correspondence between two arbitrary scenes via efficient decoupled attention.
Next, we propose an end-to-end network for joint global and local feature alignment and synthesis. Finally, we propose a novel self-supervision task to enable training. Experiments on the large-scale and more diverse COCO-stuff dataset show significant improvements over the existing methods. Moreover, our approach provides interpretability and can be readily extended to other content manipulation tasks including style and spatial interpolation or extrapolation.
\vspace{-2mm}
\end{abstract}

\section{Introduction}
\vspace{-1mm}
Conditional generative adversarial network (cGAN)~\cite{conditionalGAN} has recently made substantial progress in realistic image synthesis. In cGAN, a generator $\hat{x}=\G(c,z)$ aims to output a realistic image $\hat{x}$ with a constraint implicitly encoded by $c$. Conversely, a discriminator $\D(x,c)$ learns such a constraint from ground-truth pairs $\langle x,c\rangle$ by predicting if $\langle \hat{x},c\rangle$ is real or generated. 

\label{sec:intro}
\begin{figure}[]
	\centering
	\vspace{-2mm}
	\includegraphics[width=1.0\linewidth]{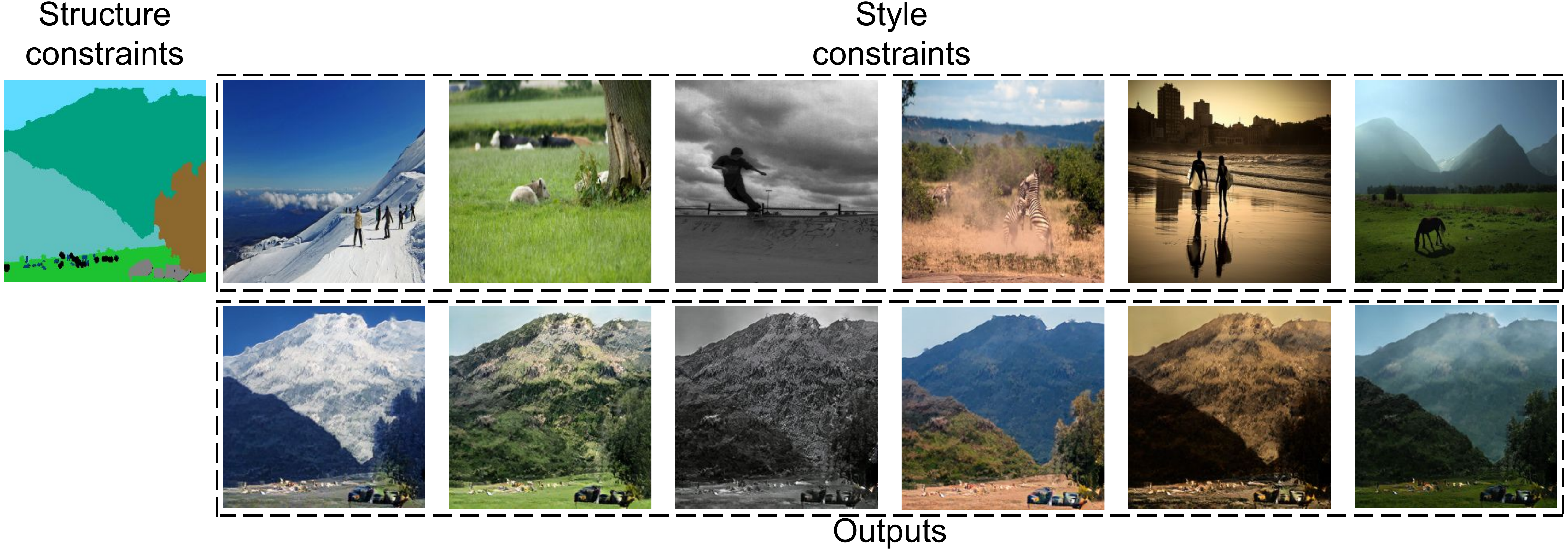}
  \setlength{\abovecaptionskip}{-0.4cm} 
	\caption{Our task aims to synthesize style-consistent images from a semantic label map (column 1) and an {\bf arbitrary} exemplar image (row 1, columns 2-7). \zht{
	In spite of the differences between the two scenes, both \emph{structurally} and \emph{semantically}, our model can synthesize high-quality images of consistent styles with the reference images.}
    }
	\label{fig:teaser}
	\vspace{-6mm}
\end{figure}

The current cGAN models~\cite{spade,pix2pixhd,pix2pix} for semantic image synthesis aim to solve the \emph{structural consistency} constraint where 
the output image $\hat{x}=\G(c)$ is required to be aligned to a semantic label map $c$.
\zht{However, for such a model, the style of $\hat{x}$ is inherently determined by the model and thus cannot be controlled by the user.}
To provide desired controllability over the generated styles, previous studies~\cite{example_cvpr18,example_cvpr19} 
\haofu{impose additional constraints and allow more inputs to the generator:} $\hat{x}_{2 \rightarrow 1} =\G(c_1, x_2, z)$,
where $x_2$ is an exemplar image that guides the style of $c_1$. However, previous studies are designed on \zht{specified} datasets such as face~\cite{liu2015deep,rossler2018faceforensics}, dancing~\cite{example_cvpr19} or street view~\cite{yu2018bdd100k}, where the exemplar images and semantic label map usually contain similar semantics and spatial structures.

\haofu{Different from the previous studies, we 
address a more challenging example-guided synthesis task that transfers styles \emph{across arbitrary scenes}. As shown in Fig.~\ref{fig:teaser}, given a semantic label map $c_1$ (column 1) and an arbitrary scene image $x_2$ (row 1, column 2-7), the task aims to generate a new scene image $\hat{x}_{2 \rightarrow 1}$ (row 2) that matches the semantic structure of $c_1$ and the scene style of $x_2$.
The challenge is that scene images have complex semantic structures as well as diversified scene styles, and more importantly, the inputs $c_1$ and $x_2$ can be \emph{structurally unaligned} and \emph{semantically different}. Therefore, a mechanism is required to better match the structures and semantics for coherent synthesis.}

\haofu{In this paper, we propose a novel Masked Spatial-Channel Attention (MSCA) module (Section~\ref{subsect:spatialchannel}) to propagate features across unstructured scenes. Our module is inspired by a recent work~\cite{doubleattention} for attention-based object recognition, \zht{but instead, we propose a new cross-attention mechanism to model the semantic correspondence for image synthesis. Moreover, our method is based on the novel design of spatial-channel decoupling that allows efficient computation.} To facilitate example-guided synthesis, we further improve the module by including: i) feature masking for semantic outlier filtering, ii) multi-scaling for global-local feature processing, and iii) resolution extending for image synthesis. As a result, our module provides both clear physical meaning and interpretability for the example-guided synthesis task.}

\haofu{We formulate the proposed approach under a unified synthesis network for joint feature extraction, alignment and image synthesis. We achieve this by applying MSCA modules to the extracted features for multi-scale feature domain alignment. Next, we apply a recent feature normalization technique, SPADE~\cite{spade} on the aligned features to allow spatially-controllable synthesis. To facilitate the learning of this network, we propose a novel self-supervision task. As opposed to~\cite{example_cvpr19}, our scheme requires only semantically parsed images for training and does not rely on video data.
We show that a model trained with this approach generalizes across different scene semantics (See Fig.~\ref{fig:teaser}).
}

Our main contributions include the following:
   \vspace{-0.08cm}
\begin{itemize}
    \item A novel masked spatial-channel attention (MSCA) module to propagate features between arbitrary scenes. 
    \vspace{-0.03cm}
    \item A unified example-guided synthesis network for joint feature extraction, alignment and image synthesis. 
    \vspace{-0.03cm}
    \item A novel self-supervision scheme that only requires semantically annotated images for training but not at the testing (image synthesis) stage. 
    \vspace{-0.03cm}
    \item Significant improvements over the existing methods on the COCO-stuff~\cite{cocostuff} dataset, as well as interpretability and easy extensions to other content manipulation tasks.
\end{itemize}

\vspace{-0.3cm}
\section{Related work}
\noindent \textbf{Generative Adversarial Networks} \quad
Recent years have witnessed the progress of generative adversarial networks (GANs)~\cite{gan} for image synthesis. A GAN model consists of a generator and a discriminator where the generator serves to produce realistic images that cannot be distinguished from the real ones by the discriminator. Recent techniques for realistic image synthesis include 
modified losses~\cite{wasserstein,lsgan,improved},
model regularization~\cite{sn},
self-attention~\cite{sagan,largescalegan}, feature normalization~\cite{stylegan} and progressive synthesis~\cite{progressivegan}.

\noindent \textbf{Image-to-Image translation (I2I)} \quad
I2I translation aims to translate images from a source domain to a target domain. The initial work of Isola~\etal~\cite{pix2pix} proposes a conditional GAN framework to learn I2I translation with paired images. Wang~\etal~\cite{pix2pixhd} improve the conditional GAN for high-resolution synthesis and content manipulation. 
To enable I2I translation without using paired data, a few works~\cite{cyclegan,liu2017unsupervised,munit,drit,pairedcyclegan} apply the cycle consistency constraint in training. Recent works on photo-realistic image synthesis take semantic label maps as inputs for image synthesis. Specifically, Wang~\etal~\cite{pix2pixhd} extend the conditional GAN for high-resolution synthesis, Chen~\etal~\cite{CRN} propose a cascade refinement pipeline. More recently, Park~\etal~\cite{spade} propose spatial-adaptive normalization for realistic image generation.

\noindent \textbf{Example-Guided Style Transfer and Synthesis} \quad
Example guided style transfer~\cite{image_analogies,Image_quilting} aims to transfer the style of an example image to a target image. Recent works~\cite{gatys2016image,adaptive_instance_normalization,phototransfer,johnson2016perceptual,deep_image_analogy,feature_shuffle,pairedcyclegan,video_style_transfer,wct2} utilize deep neural network features to model and transfer styles. Several frameworks~\cite{munit,huang2018multimodal,example_iclr19} perform style transfer via image domain style and content disentanglement. In addition, domain adaptation~\cite{pairedcyclegan} applies a cycle consistency loss to cross-domain style transformation.

More recently, example-guided synthesis~\cite{example_cvpr18,example_cvpr19} is proposed to transfer the style of an example image to a target condition, e.g. a semantic label map. Specifically, Lin~\etal~\cite{example_cvpr18} apply dual learning to disentangle the style for guided synthesis, Wang~\etal~\cite{example_cvpr19} extract style-consistent data pairs from videos for model training. 
In addition, Park~\etal~\cite{spade} adopt an I2I network under the auto-encoding framework for example-guided image synthesis.
Different from~\cite{example_cvpr18,example_cvpr19,spade}, we address style alignment issue between arbitrary scenes for {\it region and semantic aware} style integration. Furthermore, our self-supervised learning scheme does not require video data and is a generalize and more challenging auto-encoding task.


\noindent \textbf{Correspondence Matching for Synthesis} \quad
Finding correspondence is critical for many synthesis tasks. For instance, Siarohin~\etal~\cite{Deformable} apply the affine transformation on reference person images to improve pose-guided person image synthesis, Wang~\etal~\cite{vid2vid} use optical flow to align frames for coherent video synthesis. However, the affine transformation and optical flow cannot adequately model the correspondences between two arbitrary scenes.

\noindent \textbf{Efficient Attention Modeling} \quad
The self-attention~\cite{wang2018non,sagan} can capture general pair-wise correspondences. However, it is computationally intensive at high-resolution. \zht{To enable fast attention computation, GCNL~\cite{GCNL} and CCCA~\cite{CCA} respectively apply Taylor series expansion and criss-cross attention to approximate self-attention. Alternatively, $A^2$-Nets~\cite{doubleattention} factorize self-attention to solve video classification tasks. Inspired by \cite{doubleattention}, we propose an attention-based module named MSCA. It is worth noting MSCA is based on cross-attention and feature masking for modeling image correspondence.}
\section{Method}
The proposed approach aims to generate scene images that align with given semantic maps. Differ from conventional semantic image synthesis methods~\cite{pix2pix,pix2pixhd,spade}, our model takes an exemplary scene as an extra input to provide more controllability over the generated scene image. Unlike existing example-based approaches~\cite{example_cvpr18,example_cvpr19}, our model addresses a more challenging case where the exemplary inputs are structurally and semantically unaligned with the given semantic map.

\begin{figure}[]
\vspace{-3mm}
	\centering
	\includegraphics[width=1.0\linewidth]{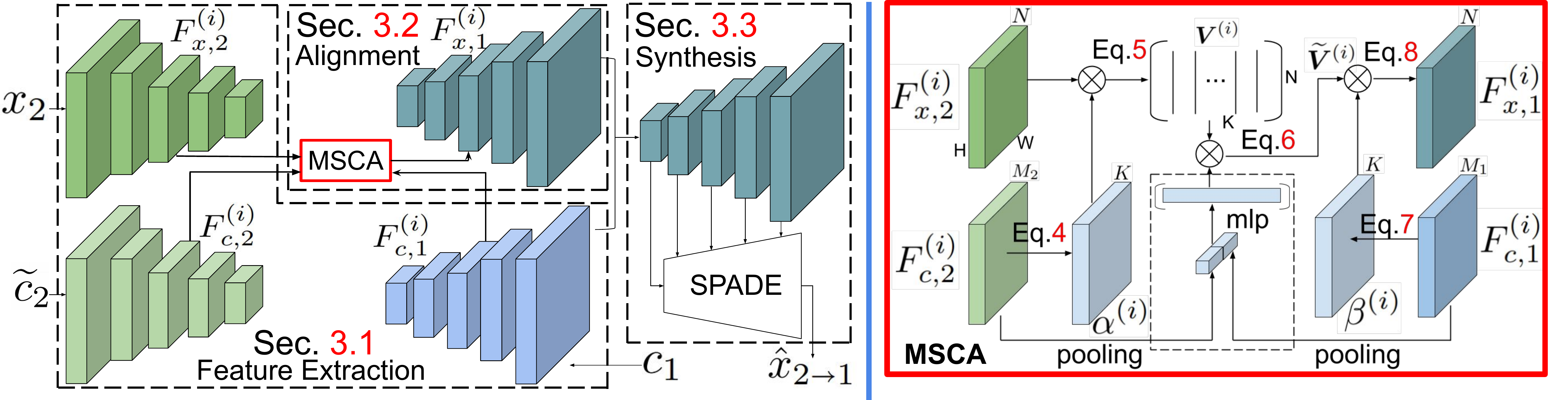}
    \setlength{\abovecaptionskip}{-0.4cm} 
	\caption{Left: the diagram of our generator. Our generator consists of three steps, namely {\it feature extraction}, {\it feature alignment}, and {\it image synthesis}. We describe each step in its corresponding section, respectively. 
	Right: The MSCA module for feature alignment at scale $i$. Our module takes image feature map $F^{(i)}_{x,2}$ and segmentation feature map $F^{(i)}_{c,1}$, $F^{(i)}_{c,2}$ as inputs to output a new image feature map $F^{(i)}_{x,1}$ that is aligned to condition $c_1$.
	}
	\label{fig:generator}
	\vspace{-4mm}
\end{figure}

Our method takes a semantic label map $c_1$, a reference image $x_2$ and its corresponding parsed semantic label map $\widetilde{c}_2$ as inputs and synthesizes an image $\hat{x}_{2\shortrightarrow 1}$ which matches the style of $x_2$ and structure of $x_1$ using a generator $\G$, $\hat{x}_{2\shortrightarrow 1}=\G(c_1,x_2,\widetilde{c}_2)$.
As shown in Fig.~\ref{fig:generator} left, the generator $\G$ consists of three parts, namely i) feature extraction ii) feature alignment and iii) image synthesis. In Sec.~\ref{subsect:encode}, we describe the first part that extracts features from inputs of both scenes. In Sec.~\ref{subsect:spatialchannel}, we propose a masked spatial-channel attention (MSCA) module to distill features and discover relations between two arbitrarily structured scenes. Unlike the affine-transformation~\cite{stn} and flow-base warping~\cite{vid2vid}, MSCA provides better interpretability to the scene alignment task. In Sec.~\ref{subsect:synthesis}, we introduce how to use the aligned features for image synthesis. Finally, in Sec.~\ref{subsect:patchsupervision}, we propose a self-supervised scheme to facilitate learning.

\subsection{Feature Extraction}
\label{subsect:encode}

Taking an image $x_2$ and label maps $c_1,\widetilde{c}_2$ as inputs, the feature extraction module extracts multi-scale feature maps for each input. Specifically, the feature map $F^{(i)}_{x,2}$ of image $x_2$ at scale $i$ is computed by:
\vspace{-1mm}
\begin{align}
\label{eq:image_feature}
\begin{aligned}
    F^{(i)}_{x,2} = W^{(i)}_x \ast F_{\text{vgg}}^{(i)}(x_2), \quad \text{for $i\in \{0, \dots, L\}$}, 
\end{aligned}
\end{align}
where $\ast$ denotes the convolution operation, $F_{\text{vgg}}^{(i)}$ denotes the feature map extracted by VGG-19~\cite{vgg} at scale $i$, and $W^{(i)}_x$ denotes a $1\times 1$ convolutional kernel for feature compression. $L$ is the number scales and we set $L=4$ in this paper.

For label map $c_1$, its feature $F^{(i)}_{c,1}$ is computed by:
\vspace{-1mm}
\begin{align}
\label{eq:segment_feature}
    F^{(i)}_{c,1}=
    \begin{cases}
        \relu(W^{(i)}_{c} \ast c^{(i)}_1) & \text{for $i=L$}, \\
        \relu(W^{(i)}_{c} \ast [\Uparrow(F^{(i+1)}_{c,1}), c_1^{(i)}]) &\text{otherwise},
    \end{cases}
\end{align}
where $\Uparrow(\cdot)$ denotes $\times 2$ bilinear interpolation, $c^{(i)}_1$ denotes the resized label map, $W^{(i)}_{c}$ denotes a $1\times 1$ convolutional kernel for feature extraction, and operation $[\cdot, \cdot]$ denotes channel-wise concatenation. Note that as scale $i$ decreases from $L$ down to $0$, the feature resolutions in Eq.~\ref{eq:segment_feature} are progressively increased to match a finer label maps $c^{(i)}_1$.

Similarly, applying Eq.~\ref{eq:segment_feature} with the same weights to label map $\widetilde{c}_2$, we can extract its features $F^{(i)}_{c,2}$:
\vspace{-1mm}
\begin{align}
\label{eq:segment_feature2}
    F^{(i)}_{c,2}=
    \begin{cases}
        \relu(W^{(i)}_{c} \ast c^{(i)}_2) & \text{for $i=L$} \\
        \relu(W^{(i)}_{c} \ast [\Uparrow(F^{(i+1)}_{c,2}), \widetilde{c}_2^{(i)}]) &\text{otherwise}
    \end{cases}.
\end{align}

\subsection{Masked Spatial-channel Attention Module}
\label{subsect:spatialchannel}


As shown in Fig.~\ref{fig:generator} right, taking the image features $F^{(i)}_{x,2}$ and the label map features $F^{(i)}_{c,1}$, $F^{(i)}_{c,2}$ as inputs\footnote{We assume spatial resolution at scale $i$ being $H\times W$ and channel size of $F^{(i)}_{x,2}$, $F^{(i)}_{c,1}$, $F^{(i)}_{c,2}$ being $N,M_1,M_2$, respectively. }, the MSCA module generates a new image feature map $F^{(i)}_{x,1}$ that has the content of $F^{(i)}_{x,2}$ but is aligned with $F^{(i)}_{c,1}$. We elaborate the detailed procedures as follows:

\noindent \textbf{Spatial Attention.} \quad
Given feature maps $F^{(i)}_{x,2}, F^{(i)}_{c,2}$ of the exemplar scene, the module first computes a spatial attention tensor $\alpha^{(i)}\in {[0,1]}^{K\cdot H\cdot W}$:
\vspace{-1mm}
\begin{align}
\label{eq:spatial-attention}
\begin{aligned}
    \alpha^{(i)} = \softmax_{2,3}(\phi^{(i)} \ast [F^{(i)}_{x,2}, F^{(i)}_{c,2}]),
\end{aligned}
\end{align}
with $\phi^{(i)}\in\mathbb{R}^{(N+M_2) \cdot K}$ denoting a $1\times1$ convolutional filter and $\softmax_{2,3}$ denoting a 2D softmax function on spatial dimensions $\{2,3\}$. The output tensor contains $K$ attention maps of resolution $H\times W$, which serve to attend $K$ different spatial regions on image feature $F^{(i)}_{x,2}$.

\noindent \textbf{Spatial Aggregation.} \quad
Then, the module aggregates $K$ feature vectors from $F^{(i)}_{x,2}$ using the $K$ spatial attention maps of $\alpha^{(i)}$ from Eq.~\ref{eq:spatial-attention}. Specifically, a matrix dot product is performed:
\vspace{-1mm}
\begin{align}
\label{eq:spatial-aggregate}
\begin{aligned}
    \matr{V}^{(i)} &= \matr{F}^{(i)}_{x,2} (\matr{\alpha}^{(i)})^\intercal,
\end{aligned}
\end{align}
with $\matr{\alpha}^{(i)}\in [0,1]^{K\cdot HW}$ and $\matr{F}^{(i)}_{x,2}\in \mathbb{R}^{C\cdot HW}$ denoting the reshaped  versions of $\alpha^{(i)}$ and $F^{(i)}_{x,2}$, respectively. The output $\matr{V}^{(i)} \in \mathbb{R}^{C \cdot K} $ stores feature vectors spatially aggregated from the $K$ independent regions of $F^{(i)}_{x,2}$.

\noindent \textbf{Feature Masking.} \quad
The exemplar scene $x_2$ may contain irrelevant semantics to the label map $c_1$, and conversely, $c_1$ may contain  semantics that are unrelated to $x_2$. 
To address this issue, we apply feature masking on the output of Eq.~\ref{eq:spatial-aggregate} by multiplying $\matr{V}^{(i)}$ with a length-$K$ gating vector at each row:
\vspace{-1mm}
\begin{align}
\label{eq:masking}
\begin{aligned}
    \widetilde{\matr{V}}^{(i)} &= (\matr{V}^{(i)})^T \circ \mlp([\gap(F^{(i)}_{c,1}),\gap(F^{(i)}_{c,2})]),
\end{aligned}
\end{align}
where $\mlp(\cdot)$ denotes a 2-layer MLP followed by a sigmoid function, $\gap$ denotes a global average pooling layer, $\circ$ denotes broadcast element-wise multiplication, and $\widetilde{\matr{V}}^{(i)}$ denotes the masked features. The design of feature masking in Eq.~\ref{eq:masking} resembles to Squeeze-and-Excitation~\cite{SEnet}. Using the integration of global information from label maps $c_1$ and $\widetilde{c}_2$, features are filtered.

\noindent \textbf{Channel Attention.} \quad
Given feature $F^{(i)}_{c,1}$ of label map $c_1$, a channel attention tensor $\beta^{(i)}\in {[0,1]}^{K\cdot H\cdot W}$ is generated as follows:
\vspace{-1mm}
\begin{align}
\label{eq:channel-attention}
\begin{aligned}
    \beta^{(i)} = \softmax_{1}(\psi^{(i)} \ast F^{(i)}_{c,1}),
\end{aligned}
\end{align}
with $\psi^{(i)}\in\mathbb{R}^{M_1\cdot K}$ denoting a $1\times1$ convolutional filter and $\softmax_{1}$ denoting a softmax function on channel dimension. The output $\beta^{(i)}$ serves to dynamically reuse features from $\widetilde{\matr{V}}^{(i)}$.

\noindent \textbf{Channel Aggregation.} \quad
With channel attention $\beta^{(i)}$ computed in Eq.~\ref{eq:channel-attention}, feature vectors at $HW$ spatial locations are aggregated again from $ \widetilde{\matr{V}}^{(i)}$ via matrix dot product:
\vspace{-1mm}
\begin{align}
\label{eq:channel-aggregate}
\begin{aligned}
    \matr{F}^{(i)}_{x,1} &=  \widetilde{\matr{V}}^{(i)} (\matr{\beta}^{(i)})^\intercal,
\end{aligned}
\end{align}
where $\matr{\beta}^{(i)}\in \mathbb{R}^{K\cdot HW}$ denotes the reshaped version of $\beta^{(i)}$. The output $\matr{F}^{(i)}_{x,1} \in \mathbb{R} ^{N \cdot HW}$ represents the aggregated features at $HW$ locations. The output feature map $F^{(i)}_{x,1}$ is generated by reshaping $\matr{F}^{(i)}_{x,1}$ to size $ N \times H \times W$.

\noindent \textbf{Remarks.} \quad
Spatial attention (Eq.~\ref{eq:spatial-attention}) and aggregation (Eq.~\ref{eq:spatial-aggregate}) attend to $K$ independent regions from feature $F^{(i)}_{x,2}$, then store the $K$ features into $\matr{V}^{(i)}$.
After feature masking, given a new label map $c_1$, channel attention (Eq.~\ref{eq:spatial-attention}) and aggregation (Eq.~\ref{eq:channel-aggregate}) combine $\widetilde{\matr{V}}^{(i)}$ at each location to compute an output feature map. As a result, each output location finds its correspondent regional features or ignored via feature masking. In this way, the feature of example scene is aligned. Note that when $K=1$ and $\alpha^{(i)}$ is constant, the above operations is essentially a global average pooling. We show in the experiment that $K=8$ is sufficient to dynamically capture visually significant scene regions for alignment.

\begin{figure}[t]
\vspace{-2mm}
	\centering
	\includegraphics[width=1.0\linewidth]{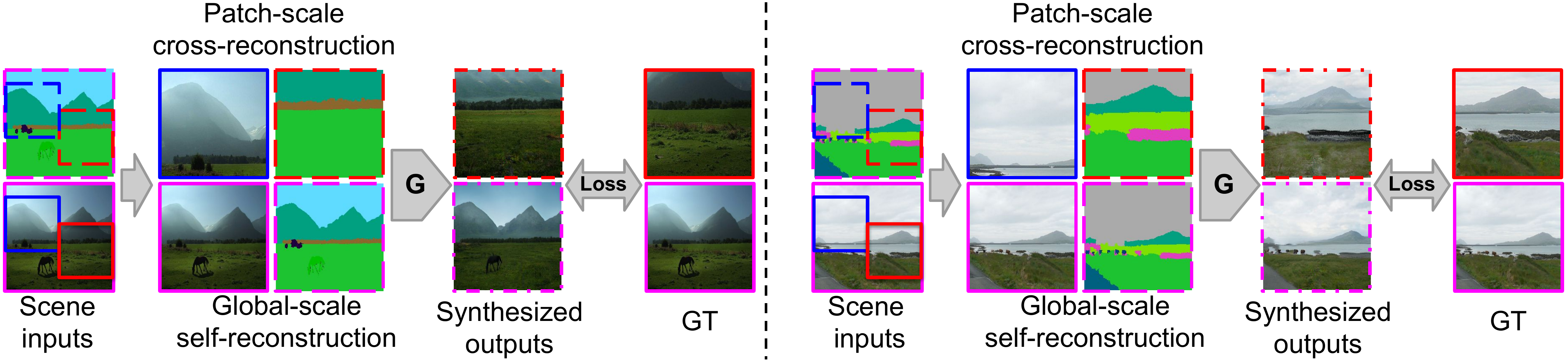}
    \setlength{\abovecaptionskip}{-0.1cm} 
	\vspace{-2mm}
	\caption{Our self-supervision scheme performs cross-reconstruction at the patch scale (top row) and self-reconstruction at the global scale (bottom row). The solid, dashed and dotted bounding boxes respective represent images, semantic label maps, and synthesized outputs. Boxes with the same color are cropped from the same position.
	} 
	\label{fig:patchsupervision}
\end{figure}

\noindent \textbf{Multi-scaling.} \quad
Both global color tone and local appearances are informative for the style-constraint synthesis. Therefore, we apply MSCA modules at all scales $i\in\{0,\dots,L\}$ to generate global and local features $F^{(i)}_{x,1}$. 

\begin{figure*}[t]
    \vspace{-1mm}
	\centering
	\includegraphics[width=1.0\linewidth]{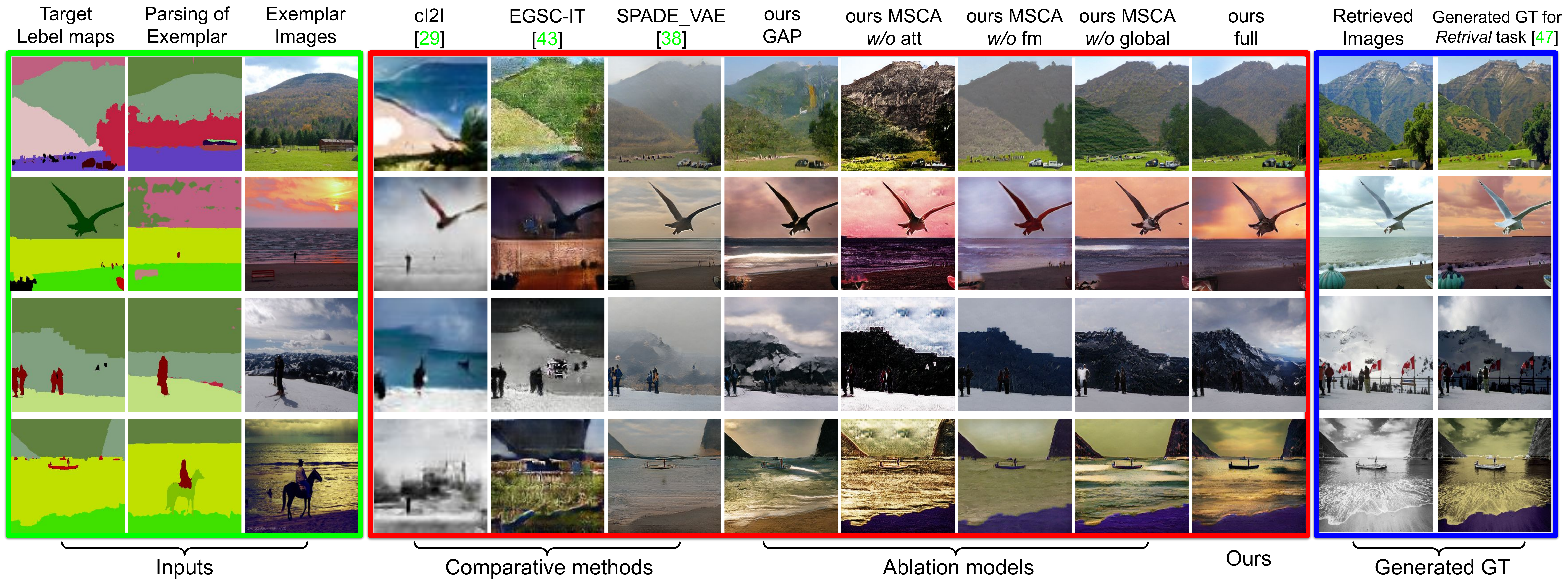}
    \setlength{\abovecaptionskip}{-0.1cm} 
	\caption{{\color{green}Green box} from left to right: the inputs for example-guided synthesis, i.e. target label maps, exemplar label parsing from Deeplab-v2~\cite{deeplabv2}, and exemplar images. {\color{red}Red box} from left to right: visual comparisons with cI2I~\cite{example_cvpr18}, EGSC-IT~\cite{example_iclr19}, SPADE\_VAE~\cite{spade}, four ablation models, and our full model. 
	{\color{blue}Blue box} from left to right: the retrieved ground-truth before and after color correction~\cite{wct2}. Our full model generates the most style-consistent results with the exemplar images.
	}
	\label{fig:compare_main}
	\vspace{-4mm}
\end{figure*}

\subsection{Image Synthesis}
\label{subsect:synthesis}
The extracted features $F^{(i)}_{c,1}$ in Sec.~\ref{subsect:encode} capture the semantic structure of $c_1$, whereas the aligned features $F^{(i)}_{x,1}$ in Sec.~\ref{subsect:spatialchannel} capture the appearance style of the example scene. In this section, we leverage $F^{(i)}_{c,1}$ and $F^{(i)}_{x,1}$ as control signals to generate output images with desired structures and styles. 

Specifically, we adopt a recent synthesis model, SPADE~\cite{spade}, and feed the concatenation of $F^{(i)}_{x,1}$ and $F^{(i)}_{c,1}$ to the spatially-adaptive denormalization layer of SPADE at each scale. By taking the style and structure signal as inputs, spatially-controllable image synthesis is achieved. We refer readers to appendix for more network details of the synthesis module.

\subsection{Self-Supervised Training}
\label{subsect:patchsupervision}
\zht{Training an example-guided synthesis model that can transfer styles across arbitrary scenes is challenging. First, style-consistent scene images are hard to acquire. A previous work~\cite{example_cvpr19} generates style-consistent pairs from videos. However, collecting scene videos can be more labor intensive. Second, even with ground truth style-consistent pairs, the trained model is not guaranteed to generalize to a new arbitrary scene.}

\zht{We propose a novel self-supervised scheme to enable style-transfer between arbitrary scenes. Our solution is motivated by the fact that the style of a scene image is stationary, meaning that patches cropped from the same scene share largely the same style. Moreover, non-overlapping patches from the same scene may contain arbitrary structures and new semantic labeling, which is essential for the learned model to generalize better.}

\zht{We first design a \emph{cross-reconstruction} task at the patch scale: given patches $x_p$ and $x_q$ cropped from the same scene image $x$, the generator is asked to reconstruct $x_p$ using $x_q$. Formally,
\vspace{-1mm}
\begin{align}
\label{eq:cross-rec}
\begin{aligned}
    \hat{x}_{p}&=\G(c_p,x_q,\widetilde{c}_q).
\end{aligned}
\end{align}
Note that $c_p$ and $\widetilde{c}_q$ contain different semantic labeling. Therefore, the generator are required to infer the correlation between different semantic labeling for coherent style transfer. An illustrative example is shown in Fig.~\ref{fig:patchsupervision}. More details on patch sampling is included in the appendix.

The cross-reconstruction task is designed at the patch scale and may not generalize well to the global scale. In fact, the generator trained with the  patch-level task alone tends to generate repetitive local textures (in Sec.~\ref{sec:experiment}). Therefore, we further design a \emph{self-reconstruction} task at the global scale, which reconstructs an global image $x$ from itself:
\vspace{-1mm}
\begin{align}
\label{eq:self-rec}
\begin{aligned}
    \hat{x}&=\G(c,x,\widetilde{c}).
\end{aligned}
\end{align}
Our training objective for generator $G$ and discriminator $D$ is formulated as:
\vspace{-1mm}
\begin{align}
\label{eq:self-rec}
\begin{aligned}
    \mathcal{L}(G, D) =& \log D(x_p,c_p,x_q,\widetilde{c}_q) + \log (1-D(\hat{x}_{p},c_p,x_q,\widetilde{c}_q)) + \mathcal{L}_{spade}(\hat{x}_{p}, x_p)\\
    &+\lambda \{\log D(x,c,x,\widetilde{c}) + \log (1-D(\hat{x},c,x,\widetilde{c})) + \mathcal{L}_{spade}(\hat{x}, x)\}
\end{aligned}
\end{align}
where $\mathcal{L}_{spade}$ refers to the VGG and GAN feature matching losses defined in~\cite{spade} and $\lambda$ is a parameter that controls the importance of the two self-supervised tasks. We set $\lambda=1$ in our experiments. Our full objective for self-supervised training is:
\vspace{-1mm}
\begin{align}
\label{eq:self-rec}
\begin{aligned}
    \min_{G^\ast} = \arg \min_{G} \max_{D} {\mathcal{L}(G, D)}
\end{aligned}
\end{align}
\vspace{-8mm}


}

\section{Experiments}
\label{sec:experiment}
\noindent \textbf{Dataset} \quad
Our model is trained on the \emph{COCO-stuff} dataset~\cite{cocostuff}. It contains densely annotated images captured from various scenes. We remove indoor images and images of random objects from the training/validation set, resulting in $34,698/499$ scene images for training/testing, respectively. 

The COCO-stuff dataset does not provide ground-truth for example-guided scene synthesis, i.e. two scene images with the exact same styles. To qualitatively evaluate the performances of example-guided synthesis model, we designed three tasks where the ground-truth image can be obtained: 
i) \emph{duplicating task} requires a model to self-reconstruct an image using its semantic label map and itself as exemplar input,
ii) \emph{mirroring task} requires a model to self-reconstruct an image using its semantic label map and the mirrored image as exemplar input,
iii) \emph{retrieving task}: requires a model to reconstruct an ground-truth (GT) image using its semantic label map and a retrieved image from a image pool. To retrieve an image that best match GT in styles, we first select 20 candidate images from the image pool that has the greatest label histogram intersections with the GT image. Afterwards, the best-matched image is select out of candidates using SIFT Flow~\cite{siftflow}. Finally, since the color of GT is different from the retrieved image, we apply color correction~\cite{wct2} on GT to eliminate color discrepency. Examples of GT (before and after color correction) are shown in the blue box in Fig.~\ref{fig:compare_main}.

\begin{table}[]
\vspace{-3mm}
	\centering
	\resizebox{\columnwidth}{!}
	{
	    \newcommand{\myhline}{\cline{2-10}}
	    \newcolumntype{L}{>{\centering\arraybackslash}m{1.7cm}}
	    \newcolumntype{X}{>{\centering\arraybackslash}m{1.22cm}}
	    \newcolumntype{M}{>{\centering\arraybackslash}m{1.4cm}}
	    
		\begin{tabular}{ |c|r|X|X|X|X|L|L|L|M|}
			\hline
		     task& measures& cI2I~\cite{example_cvpr18}&	EGSC-IT~\cite{example_cvpr19} &  SPADE-VAE~\cite{spade}	& ours GAP & \makecell{ ours MSCA \\w/o att } & \makecell{ours MSCA \\w/o fm} & \makecell{ours MSCA\\w/o global} & ours full \\
		     \hline
		    \multirow{ 4}{*}{\em retrieving} &PSNR$\uparrow$ & 9.50& 12.57 & 15.77 & 15.85& 11.96 & 16.24 & 15.98 &\best{16.65}\\
		     \myhline
		    &LPIPS$\downarrow$ & 0.757 & 0.581& 0.483& 0.457& 0.522& 0.451& 0.446& \best{0.437}\\
		     \myhline
		    &FID$\downarrow$ & 228.63&163.23&102.68&101.74&112.83&100.01&96.66&\best{91.91}\\
		     \myhline
		    &$\mathcal{L}_{style}\downarrow$ & 3.53e-3 & 1.69e-3 & 1.07e-3 & 6.40e-4& 7.10e-3& 7.62e-4 & 7.21e-4 & \best{5.34e-4}\\
			\myhline
		     \hline
		    \multirow{ 4}{*}{\em mirroring} &PSNR$\uparrow$ & 9.46& 12.44 & 15.37 & 15.80& 11.95 & 16.02 & 16.58 &\best{17.03}\\
		     \myhline
		    &LPIPS$\downarrow$ & 0.759 & 0.602& 0.477& 0.438& 0.510& 0.437& 0.421& \best{0.397}\\
		     \myhline
		    &FID$\downarrow$ & 242.73&190.01&90.99&89.15&102.41&90.52&85.92&\best{76.75}\\
		     \myhline
		    &$\mathcal{L}_{style}\downarrow$ & 4.14e-3 & 2.03e-3 & 1.67e-3 & 7.45e-4& 7.99e-3& 8.76e-4 & 6.69e-4 & \best{3.96e-4}\\
		     \hline
		    \multirow{ 4}{*}{\em duplicating} &PSNR$\uparrow$ & 9.46& 12.45 & 15.43 & 15.80& 11.94 & 16.02 & 16.62 &\best{17.03}\\
		     \myhline
		    &LPIPS$\downarrow$ & 0.759 & 0.602& 0.476& 0.438& 0.510& 0.438& 0.421& \best{0.397}\\
		     \myhline
		    &FID$\downarrow$ & 242.81&190.02&90.97&89.22&103.23&90.56&86.20&\best{76.64}\\
		     \myhline
		    &$\mathcal{L}_{style}\downarrow$ & 4.15e-3 & 2.03e-3 & 1.66e-3 & 7.41e-4& 7.94e-3& 8.81e-4 & 6.53e-4 & \best{3.97e-4}\\
			\hline
		\end{tabular}
	}
	\caption{Quantitative comparisons of different methods and ablation models in terms of PSNR, LPIPS~\cite{lpips}, Fréchet Inception Distance (FID)~\cite{fid} and style loss ($\mathcal{L}_{style}$)~\cite{gatys2015neural}. Higher scores are better for metrics with uparrow ($\uparrow$), and vice versa.
	}
	\label{tab:main_experiment}
	\vspace{-5mm}
\end{table}

\begin{figure*}[t]
\vspace{-2mm}
	\centering
	\includegraphics[width=1.00\linewidth]{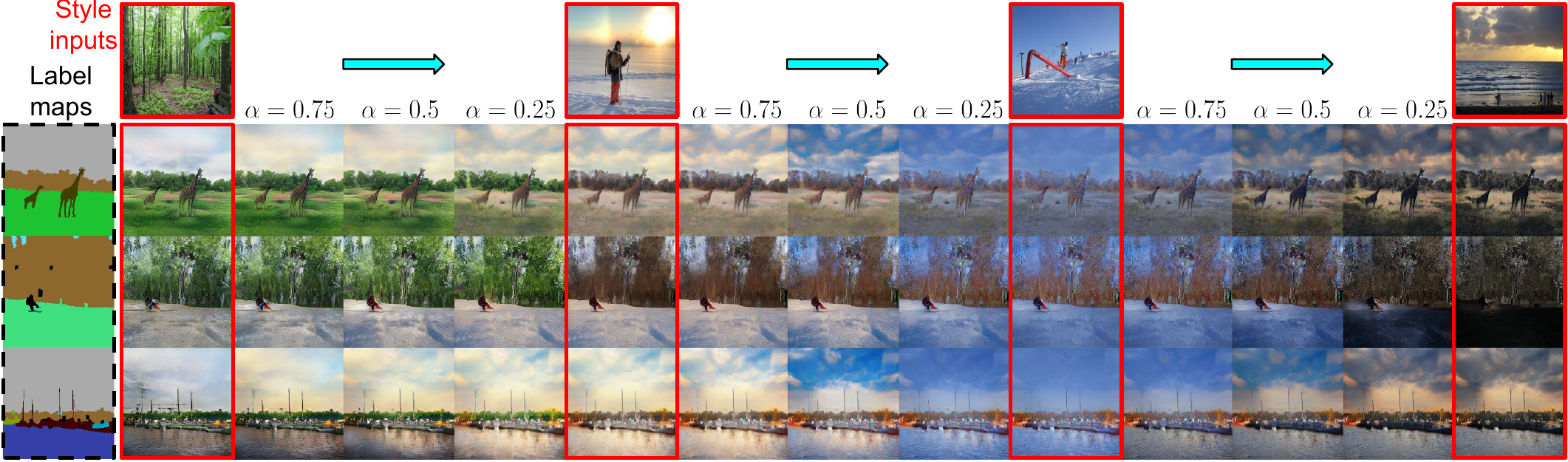}
    \setlength{\abovecaptionskip}{-0.4cm} 
	\caption{With a slight modification and no further training, our model can perform style interpolation between exemplar inputs. 
	Note that our model can interpolate styles for new semantics, e.g. ``river'' in row 3.
	Please refer to Interpolation, Sec.~\ref{sec:experiment} for details.
    }
	\label{fig:interpolate}
	\vspace{-4mm}
\end{figure*}

\noindent \textbf{Implementation Details} \quad
We use a COCO-stuff pretrained Deeplab-v2~\cite{deeplabv2} model to generate semantic label maps from exemplar images. During training, we resize images to $512 \times 512$ then crop two non-overlapping patches of size $256 \times 256$ to facilitate patch-based cross-reconstruction. After $20$ epochs, we increase the patch size to $384 \times 384$ for cross-patch reconstruction in order to improve generalization to global scenes. Details of the patch sampling procedure are provided in the appendix.

For the MSCA modules from scale $0$ to $4$, the number of attention maps $K$ are respectively set to $8,16,16,16,16$. The learning rate is set to $0.0002$ for the generator and the discriminator. The weights of generator are updated every $5$ iterations. We adopt the Adam~\cite{adam} optimizer ($\beta_1=0.9$ and $\beta_2=0.999$) in all experiments. Our synthesis model and all comparative models based on SPADE backbone are trained for $40$ epochs to generate the results in the experiments.

Before training, we pretrain the spatial-channel attention with a lightweight feature decoder to avoid the backpropagatation through the extremely heavy SPADE model. Specifically, at each scale, the concatenation of $F^{(i)}_{x,1}$ and $F^{(i)}_{c,1}$ in Sec.~\ref{subsect:synthesis} at each scale is fed into a $1\times 1$ convolutional layer to reconstruct the ground-truth VGG feature at the corresponding scale. The pretraining takes around ~$4$\% of the total training time to converge. More details of the pretraining procedure is provided in the appendix.

\noindent \textbf{Comparative Methods} \quad
We compare our approach with an example-guided synthesis approach: variational autoencoding SPADE (\texttt{SPADE\_VAE})~\cite{spade} which is based on a  self-reconstruction loss for training.
We also trained \texttt{cI2I}~\cite{example_cvpr18}, \texttt{EGSC-IT}~\cite{example_iclr19} and \texttt{SCGAN}~\cite{example_cvpr19} on COCO-stuff dataset.
\texttt{cI2I} and \texttt{EGSC-IT} are originally designed for exemplar-guided image-to-image translation. As a result, we observed that \texttt{cI2I} and \texttt{EGSC-IT} have difficulty generating images from one-hot encoded semantic label maps. However, these models can synthesize reasonable images from color-encoded semantic label maps.
Finally, we note that \texttt{SCGAN} is not directly applicable to COCO-stuff dataset, as its positive pairs are sampled from video data. We attempted to modify \texttt{SCGAN} such that its positive pairs can be generated from our self-supervision task. However, we could not achieve reasonable image outputs. We speculate that the negative sampling and semantic consistency loss of \texttt{SCGAN} is not optimal for COCO-stuff dataset, as 
COCO-stuff dataset contains much larger variations for negative pairs. 
Finally, four ablation models are evaluated (see Ablation Study).

\noindent \textbf{Quantitative Evaluation} \quad
For quantitative evaluation, we apply PSNR as the low-level metric.
Furthermore, perceptual-level metrics including Perceptual Image Patch Similarity Distance (LPIPS)~\cite{lpips}, Fréchet Inception Distance (FID)~\cite{fid} and style loss ($\mathcal{L}_{style}$) of~\cite{gatys2015neural} are evaluated on different methods. The linearly calibrated VGG model is used to compute LPIPS distance.

Among the four competitive methods (\texttt{cI2I}, \texttt{EGSC-IT}, \texttt{SPADE\_VAE} and \texttt{ours full}) in Table \ref{tab:main_experiment}, our method clearly outperforms the remaining methods both in low-level and perceptual-level measurements, suggesting that our model can better preserve color and texture appearances.
Also, we observe that without further modification, the off-the-shelf example-guided image translation approaches cannot perform well on image synthesis tasks (\texttt{cI2I}, \texttt{EGSC-IT}). It suggests that example-guided image-synthesis task can be more challenging. Finally, a simple synthesis model (\texttt{ours GAP}) outperforms SPADE\_VAE, suggesting that the self-supervised task in Sec.~\ref{subsect:patchsupervision} is beneficial to the exampled-guided synthesis task (see Ablation Study for more details).


\begin{figure*}[t]
\vspace{-2mm}
	\centering
	\includegraphics[width=1.0\linewidth]{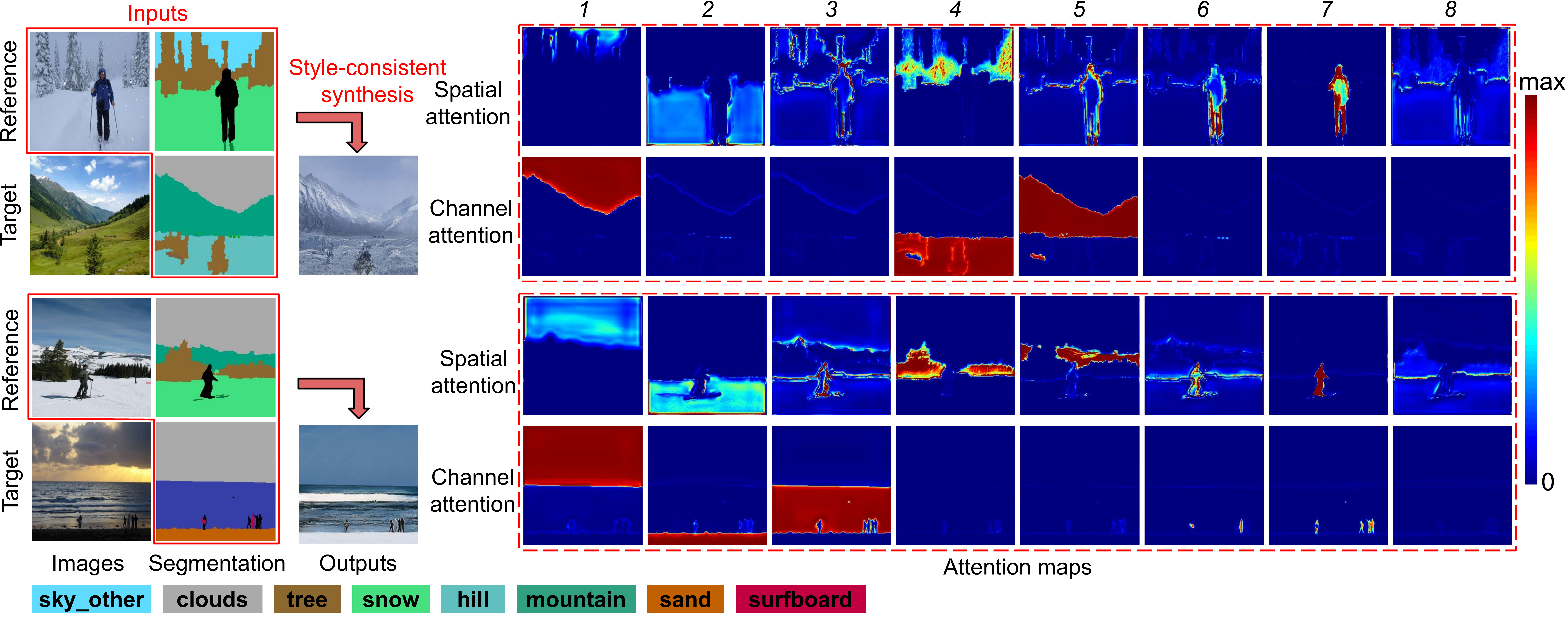}
  \setlength{\abovecaptionskip}{-0.4cm} 
	\caption{Left: inputs and outputs of our model. Right: the $K=8$ learned spatial and channel attention that attends and transfer feature between individual exemplar and target regions. By examining the semantics label maps, we observe the following transformation patterns: $\texttt{sky\_other} \shortrightarrow \texttt{clouds}$, $\texttt{tree} \shortrightarrow \texttt{\{tree, hill\}}$ for the 1st sample, and $\texttt{clouds} \shortrightarrow \texttt{clouds}$, $\texttt{snow} \shortrightarrow \texttt{sand}$, $\texttt{other} \shortrightarrow \texttt{\{surfboard,other\}}$ for the 2nd sample.
	}
	\label{fig:attention}
	\vspace{-4mm}
\end{figure*}

\noindent \textbf{Qualitative Evaluation} \quad
Fig.~\ref{fig:compare_main} qualitatively compares our approach against the remaining approach on four scenes.
We observe that our full model generates more style-consistent results with the exemplar images. In comparisons, SPADE\_VAE tends to generate results with low color contrast, as it lacks the mechanism and supervision to perform region-aware style transformation. In addition, the existing example-guided image-to-image approaches (\texttt{cI2I}, \texttt{EGSC-IT}) cannot generalize well to the image synthesis tasks.



\noindent \textbf{Ablation Study} \quad
To evaluate the effectiveness of our design, we separately train four variants of our model: i) \texttt{our GAP} that replaces the MSCA module with global average pooling,
ii) \texttt{ours MSCA w/o att} that keeps MSCA modules but replaces spatial and channel attention with one-hot label maps from source and target domains, respectively. In such a way, alignment is performed only for regions with the same semantic labeling, iii) \texttt{ours MSCA w/o fm} that keeps MSCA modules but removes the feature masking procedures, and iv) \texttt{ours MSCA w/o global} that is trained without using global-level self-reconstruction (Eq.~\ref{eq:self-rec}) or increased patchsize.

\zht{
In Table~\ref{tab:main_experiment}, our full model clearly achieves the best qualitative results. 
In Fig.~\ref{fig:compare_main}, \texttt{ours GAP} tends to produce images with deviated colors since it averages the style features from all exemplar regions. In contrast, our model dynamically transfers appearance for individual regions. We observe that \texttt{ours w/o att} is less stable in training and cannot generate plausible results. We suspect that the label-level alignment generates more misaligned and noisier feature maps, thus hurting training. \texttt{ours MSCA w/o fm} tends to generate inconsistent  colors for new semantic labels, for instance, the ``hill'' and ``sky'' regions in rows 1 and 2 of Fig.~\ref{fig:compare_main}. In contrast, our model can eliminate the undesired influence of exemplar inputs on new semantic labels. \texttt{ours MSCA w/o global} performs reasonably well but it tends to generate repetitive local textures, while the self-reconstruction scheme helps our model  generalize better at the global scale.
}

\noindent \textbf{User Study} \quad
\zht{
We conduct a user study to qualitatively evaluate our method. Specifically, we retrieve an exemplar image for each testing label map, and ask 20 subjects to choose the most style-consistent results generated by our method and two competitive baselines (\texttt{SPADE\_VAE} and \texttt{ours GAP}). To generate samples for the  user study, we first rank the label histogram intersections with each target scene for all images in the image pool, and use the top 20 percentile images as exemplars\footnote{This differs from the typical retrieve task that uses the top-1 image since the top 20 percentile images tend to be  more semantically different from the target label maps.}. The subjects are given unlimited time to make their selections. For each subject, we randomly generate 100 questions from the dataset. Table~\ref{tab:user_study} shows the evaluation results. First, all subjects strongly favor our results. Second, \texttt{ours GAP} is favored more than twice over \texttt{SPADE\_VAE}~\cite{spade}, further suggesting that the proposed self-supervision scheme is effective since \texttt{ours GAP} is also trained with self-supervision. 
\begin{table}[]
\vspace{-3mm}
	\centering
    \newcolumntype{L}{>{\centering\arraybackslash}m{2.5cm}}
	\begin{tabular}{ |c|L|L|L|}
		\hline
		Methods   &	SPADE\_VAE~\cite{spade} & ours GAP &  ours full\\
		\hline
		Choose rate     & 15.6       & 29.3     &   55.0    \\
		\hline
	\end{tabular}
	\label{tab:user_study}
	\caption{User preference study. The numbers indicate the percentage of user who favors the result generated by different methods. Two com}
	\vspace{-4mm}
\end{table}
}

\noindent \textbf{Effect of Attention} \quad
To understand the effect of spatial-channel attention, we visualize the learned spatial and channel attention in Fig.~\ref{fig:attention}. We observe that: 
a) spatial attention can attend to multiple regions of the reference image. For each reference region, channel attention finds the corresponding target region.
b) spatial-channel attention can detect and utilize the similarities of semantic labels to facilitate style features transfer. In the first sample of Fig.~\ref{fig:attention}, attention in channels $1,4$ respectively perform transformations: $\texttt{sky\_other} \shortrightarrow \texttt{clouds}$, $\texttt{tree} \shortrightarrow \texttt{\{tree, hill\}}$. In the second sample, attention in channels $1,2,7$ respectively perform transformations: $\texttt{clouds} \shortrightarrow \texttt{clouds}$, $\texttt{snow} \shortrightarrow \texttt{sand}$ and $\texttt{other} \shortrightarrow \texttt{\{surfboard,other\}}$. We provide more analysis on the effect of attention in the appendix.


\begin{figure}[t]
\vspace{-2mm}
	\centering
	\includegraphics[width=0.6\linewidth]{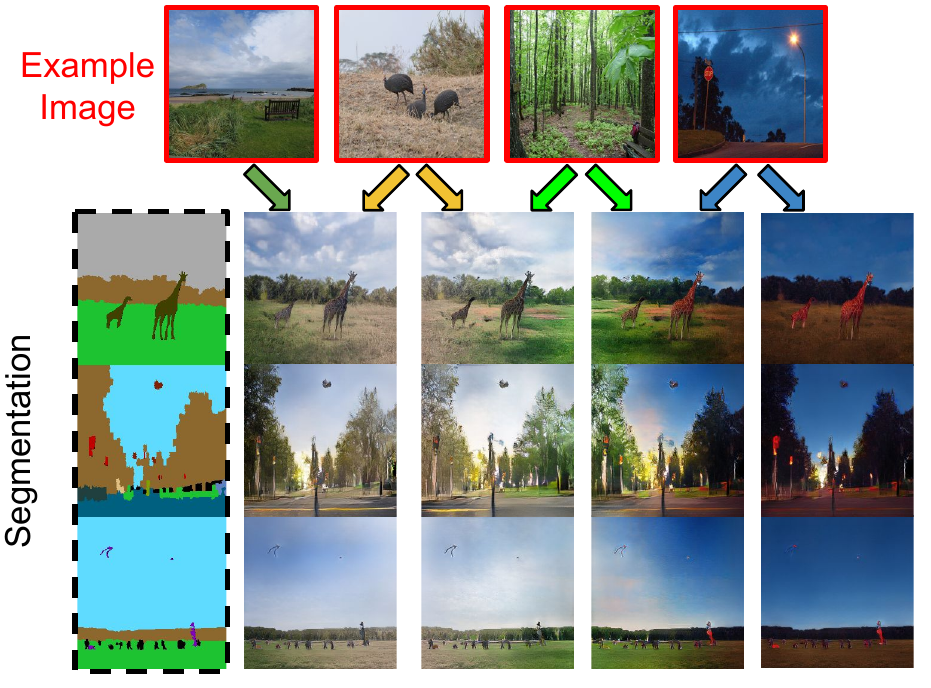}
  \setlength{\abovecaptionskip}{-0.1cm} 
	\caption{
	With a slight modification and no further training, our model can perform spatial style interpolation. In this figure, we demonstrate a horizontal gradient style change on the output image. Please refer to Interpolation, Sec. 4 for more details.
	}
	\label{fig:interpolate_spatial}
	\vspace{-4mm}
\end{figure}

\noindent \textbf{Interpolation} \quad
We can easily control the synthesized styles in the test stage by manipulating attentions. Here, we show how to interpolate between two styles using our trained model: given two example images $x_2$ and $x_3$, we first compute their image features $F^{(i)}_{x,2},F^{(i)}_{x,3}$ and the spatial-attention maps $\alpha^{(i)}_2, \alpha^{(i)}_3$. Given an interpolating factor $\alpha\in[0,1]$ where $\alpha=1$ means ignoring the example scene $x_3$, the spatial attention map of the first scene is modified by $\alpha^{(i)}_2 \coloneqq \alpha^{(i)}_2+\log(\frac{\alpha^{(i)}_2}{1-\alpha^{(i)}_2})$. Afterwards, both feature maps $F^{(i)}_{x,2},F^{(i)}_{x,3}$ and spatial attention $\alpha^{(i)}_2, \alpha^{(i)}_3$ are concatenated along the horizontal axis. In addition, the masking score (output of the 2-layer MLP in Eq.~\ref{eq:masking}) is also interpolated. With the remaining procedures unchanged, i.e., same spatial aggregation, feature masking, channel aggregation and synthesis, interpolation results are readily generated. As shown in Fig.~\ref{fig:interpolate}, with slight modifications, our model can perform effective style interpolation. Specifically, the style traverses for four distinctive exemplar styles are achieved in Fig.~\ref{fig:interpolate}.

Likewise, by manipulating the channel attention at each spatial location, it is possible to adaptively mix style to synthesize an output image, i.e. spatial styles interpolation. As shown in Figure~\ref{fig:interpolate_spatial}, using the previous input, we interpolate between styles from left to right in a single image.

\noindent \textbf{Extrapolation} \quad 
\begin{figure}[t]
	\centering
	\includegraphics[width=0.98\linewidth]{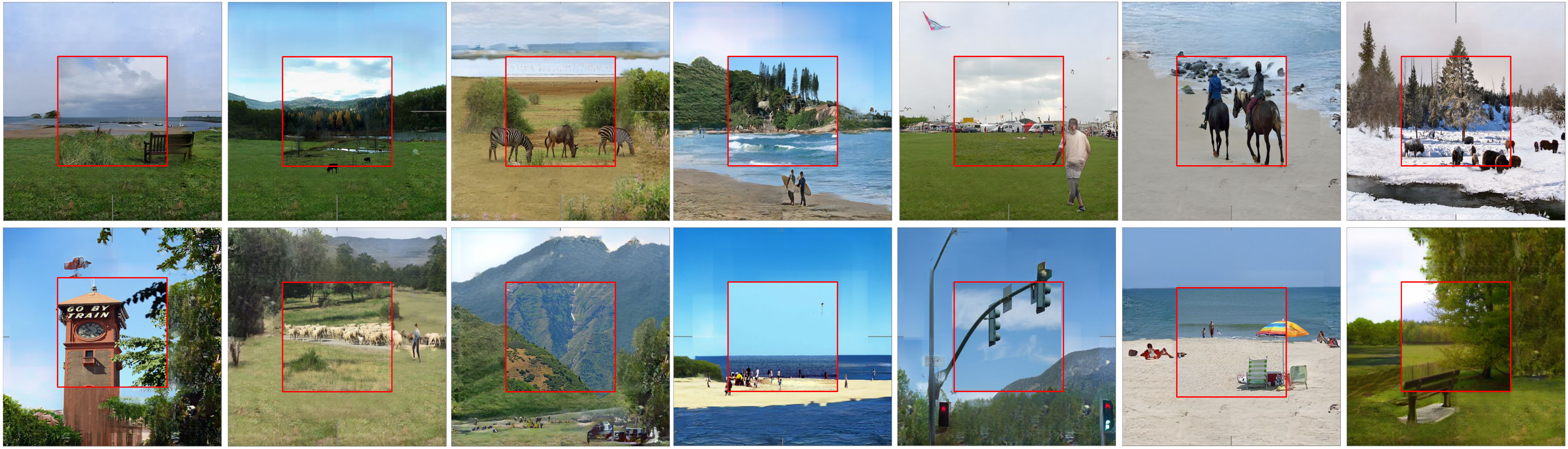}
  \setlength{\abovecaptionskip}{-0.1cm} 
	\caption{Given an exemplar patch at the center and the global semantic label map, our model can perform example-guided scene image extrapolation, i.e.  generating style-consistent beyond-the-border images content guided by semantic maps.}
	\label{fig:extrapolation}
\end{figure}
\vspace{-2mm}
Given a scene patch at the center our model can achieve scene extrapolation, i.e.  generating beyond-the-border image content according to the semantic map guidance. A $512\times 512$ extrapolated image is generated by weighted combining synthesized $256\times256$ patches at $4$ corners and $10$ other random locations. As shown in Fig.~\ref{fig:extrapolation}, our model generates visually plausible extrapolated images, showing the promise of our proposed framework for guided scene panorama generation.

\begin{figure}[t]
	\centering
	\includegraphics[width=0.55\linewidth]{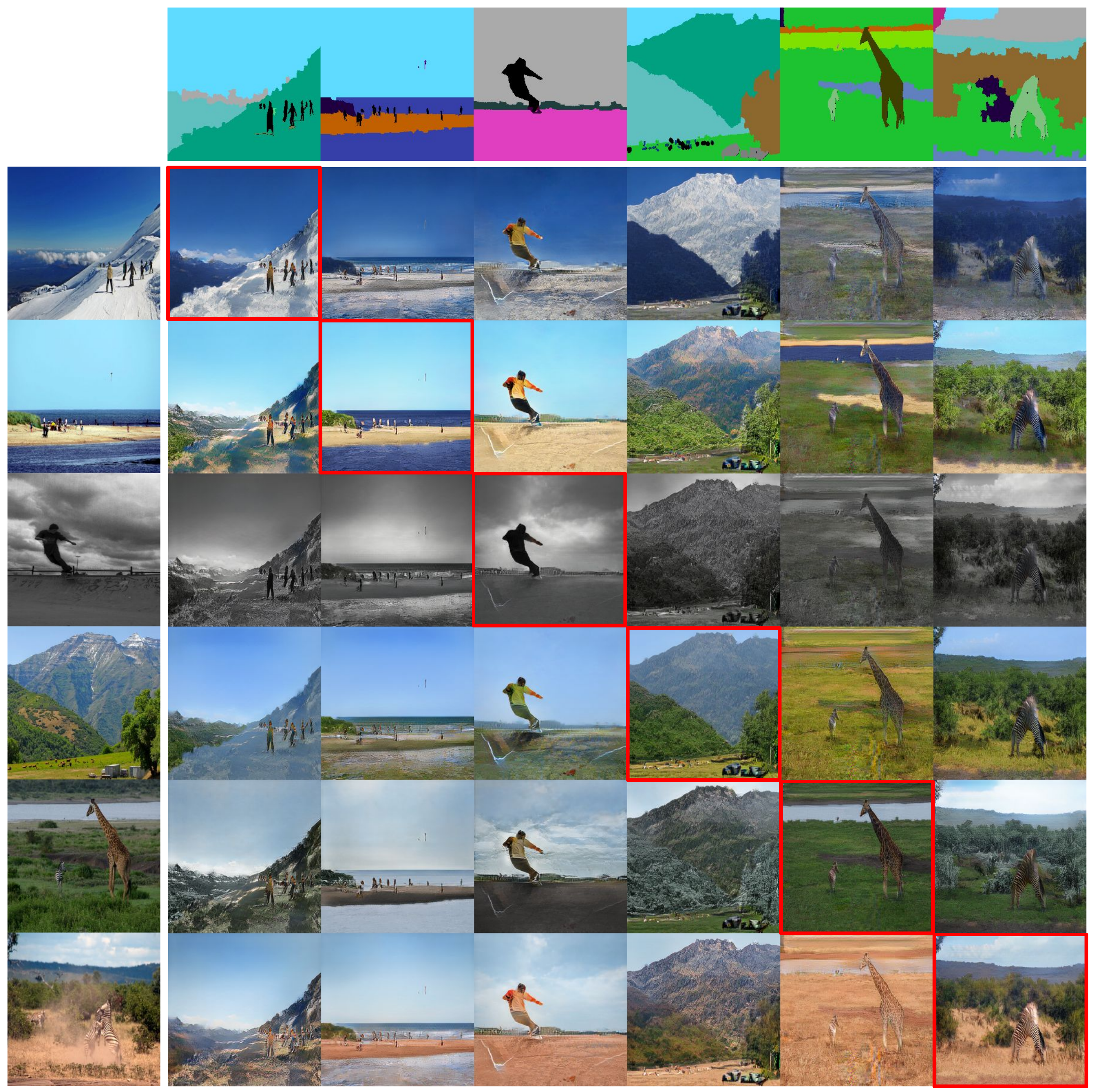}
  \setlength{\abovecaptionskip}{-0.1cm} 
	\caption{Style-structure swapping on 6 arbitrary scenes at resolution $256\times 256$.
    Our model can generalize across recognizably different scenes of different semantics, and synthesize images with reasonable and consistent styles. Note that the images along the diagonal (red boxes) are {\it self-reconstruction}. Therefore, the more they resemble the source images in column 1, the better the algorithm is. Please zoom in for details.
    }
	\label{fig:swap1}
\end{figure}

\noindent \textbf{Style Swapping} \quad
Fig.~\ref{fig:swap1} shows reference-guided style swapping between six arbitrary scenes. Our model can generalize across recognizably different scenes semantics, including snow mountain, seashore, urban, mountain, grassland and dessert, and synthesize image with reasonable and consistent styles. More results and comparisons to other approaches are included in the appendix.

\section{Conclusion}
We propose to address a challenging example-guided image synthesis task between arbitrary scenes. To propagate information between two structurally unaligned and semantically different scenes, we propose an MSCA module that leverages decoupled cross-attention for adaptive correspondence modeling. With MSCA, we propose a unified model for joint global-local alignment and image synthesis. We further propose a patch-based self-supervision scheme that enables training. Experiments on the COCO-stuff dataset show significant improvements over the existing methods. Furthermore, our approach provides interpretability and can be extended to other content manipulation tasks.

\clearpage
%
%
\bibliographystyle{splncs04}
\bibliography{egbib}

\newpage
\appendix
\section{Appendix}
In this appendix, we provide more analysis of the learned attention in Sec.~\ref{sec:attention}, results on an additional dataset in Sec.~\ref{sec:indoor_dataset}, more results of scenes swapping in Sec.~\ref{sec:swap}, more details of the synthesis module in Sec~\ref{sec:synthesis}, patch sampling strategy and more training details in Sec.~\ref{sec:patch}, pretraining of MSCA modules in Sec.~\ref{sec:pretraining}.

\vspace{-3px}
\section{Analysis of the Learned Attention}
\label{sec:attention}
\begin{figure}[]
	\centering
	\vspace{-2mm}
	\includegraphics[width=1.0\linewidth]{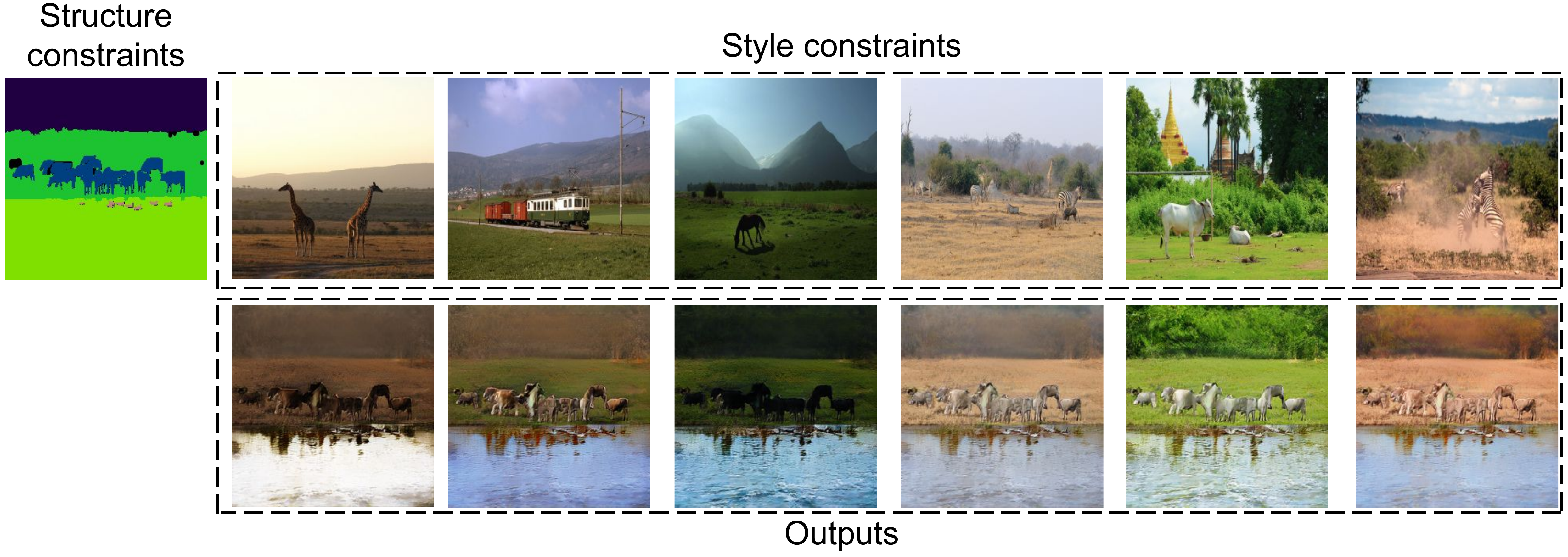}
  \setlength{\abovecaptionskip}{-0.4cm} 
	\caption{Our model can transfer appearance from \texttt{sky} to \texttt{river} for better example-guided image synthesis.}
	\label{fig:teaser2}
	\vspace{-6mm}
\end{figure}

As shown in Fig.~\ref{fig:teaser2}, our model is able to utilize the appearance of \texttt{sky} from reference images to better synthesize the appearance of \texttt{river}, showing that our learned attention can associate correlation of appearance for better example-guided image synthesis.

\begin{figure*}[h]
	\centering
	\includegraphics[width=0.7\linewidth]{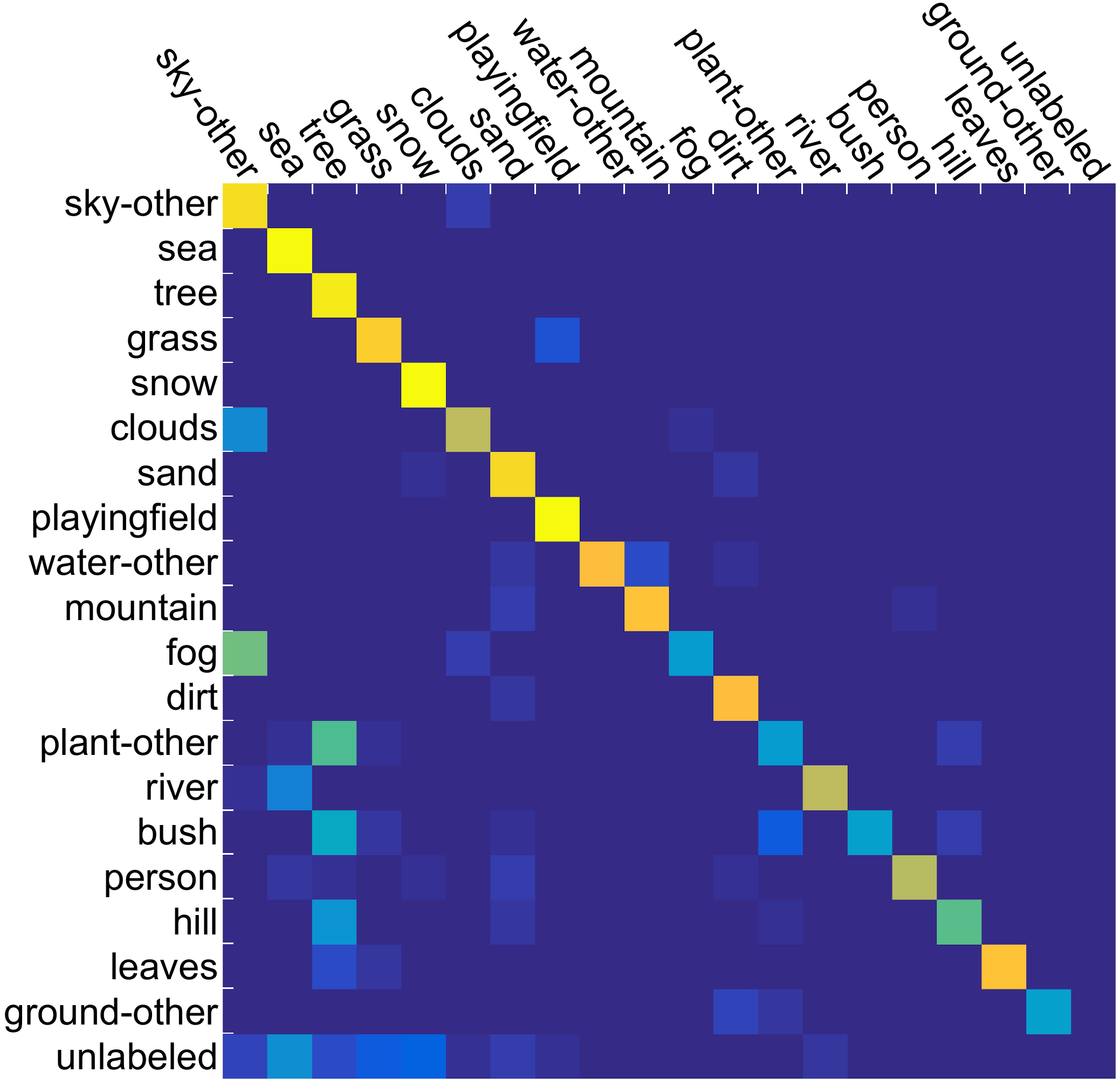}
	\caption{The transfer matrix visualizes how our learned attention transform features from semantics of exemplar images (vertical axis) to semantics of target label maps (horizontal axis) on the test set. Colors from blue to red represent ratios from 0 to 1. The most appeared $20$ semantics are shown in figure. The learned attention can transfer across semantic labels with similar appearance, for instance, from $\{\texttt{plant-other,bush,hill,leaves}\}$ to \texttt{tree} on column $3$.
	}
	\label{fig:transfer_matrix}
\end{figure*}

To better understand how the learned attention transfer features across different semantics from exemplar images to target label maps, the following visualization is performed on test set. Specifically, we use the learned MSCA at the finest scale to transfer one-hot encoded semantic label maps of the exemplar images, then compare the transferred results against the target semantic label map. The resulting transfer matrix is visualized in Fig.~\ref{fig:transfer_matrix}. From the figure, the learned attention tends to transfer feature across semantically similar labels, such as $\{\texttt{plant-other,bush,hill,leaves}\}\shortrightarrow \texttt{tree}$, $\{\texttt{clouds,fog}\}\shortrightarrow \texttt{sky-other}$,
$\{\texttt{grass}\}\shortrightarrow \texttt{playingfield}$ and $\{\texttt{bush}\}\shortrightarrow \texttt{plant-other}$. It shows that our attention can automatically discover semantics labels that shares similar visual appearance. 

\label{sec:intro}

\begin{figure*}[h]
	\centering
	\includegraphics[width=1\linewidth]{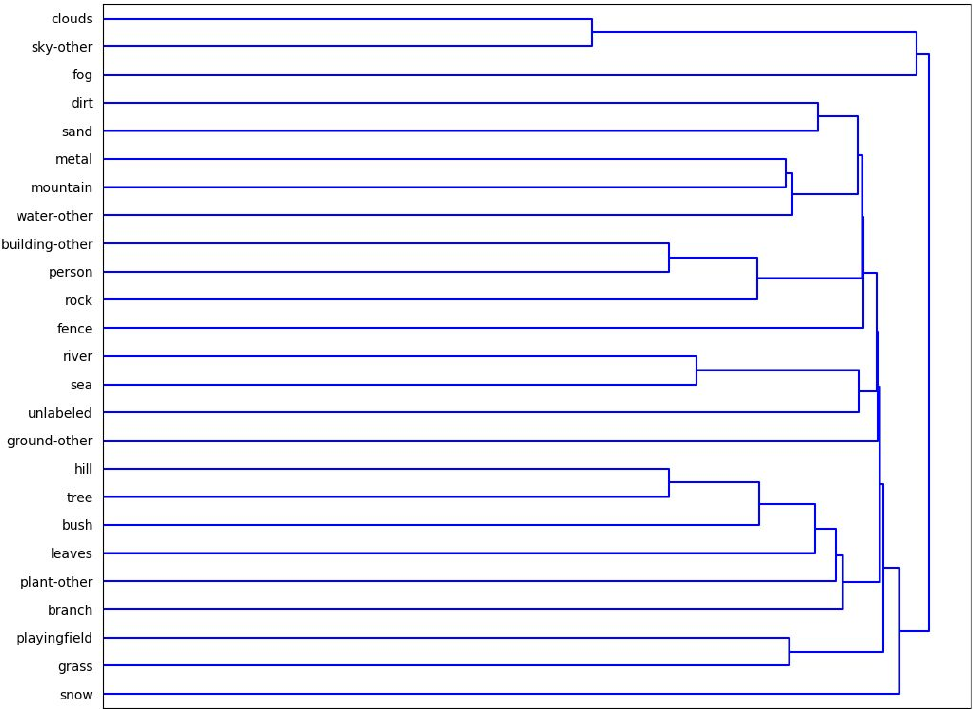}
	\caption{Hierarchical clustering on semantics by using each row of transfer matrix (Fig.~\ref{fig:transfer_matrix}) as feature vector (See Sec.~\ref{sec:attention}). Our learned attention can discovery similar semantics such as $\{\texttt{clound},\texttt{sky-other}\}$, $\{\texttt{river},\texttt{sea}\}$, $\{\texttt{dirt},\texttt{sand}\}$ without additional semantic-level supervision.
	}
	\label{fig:dendrogram}
\end{figure*}

In addition, we perform a hierarchical clustering on the top-$25$ semantic labels on the test set. Specifically, we use each row of the transfer matrix as the feature vector of each semantic labels. The diagonal of the transfer matrix is set to the second largest value of each row. From the dengrogram visualized in Fig.~\ref{fig:dendrogram}, we observe that semantics with similar appearances such as $\{\texttt{cloud},\texttt{sky-other}\}$, $\{\texttt{river},\texttt{sea}\}$, $\{\texttt{dirt},\texttt{sand}\}$ tend to group together, suggesting that our attention model can discover semantic labels with similar visual appearances for effective feature transfer.

\vspace{-3mm}
\section{Results on an Additional Dataset}
\label{sec:indoor_dataset}
\begin{figure*}[]
	\centering
	\includegraphics[width=1.0\linewidth]{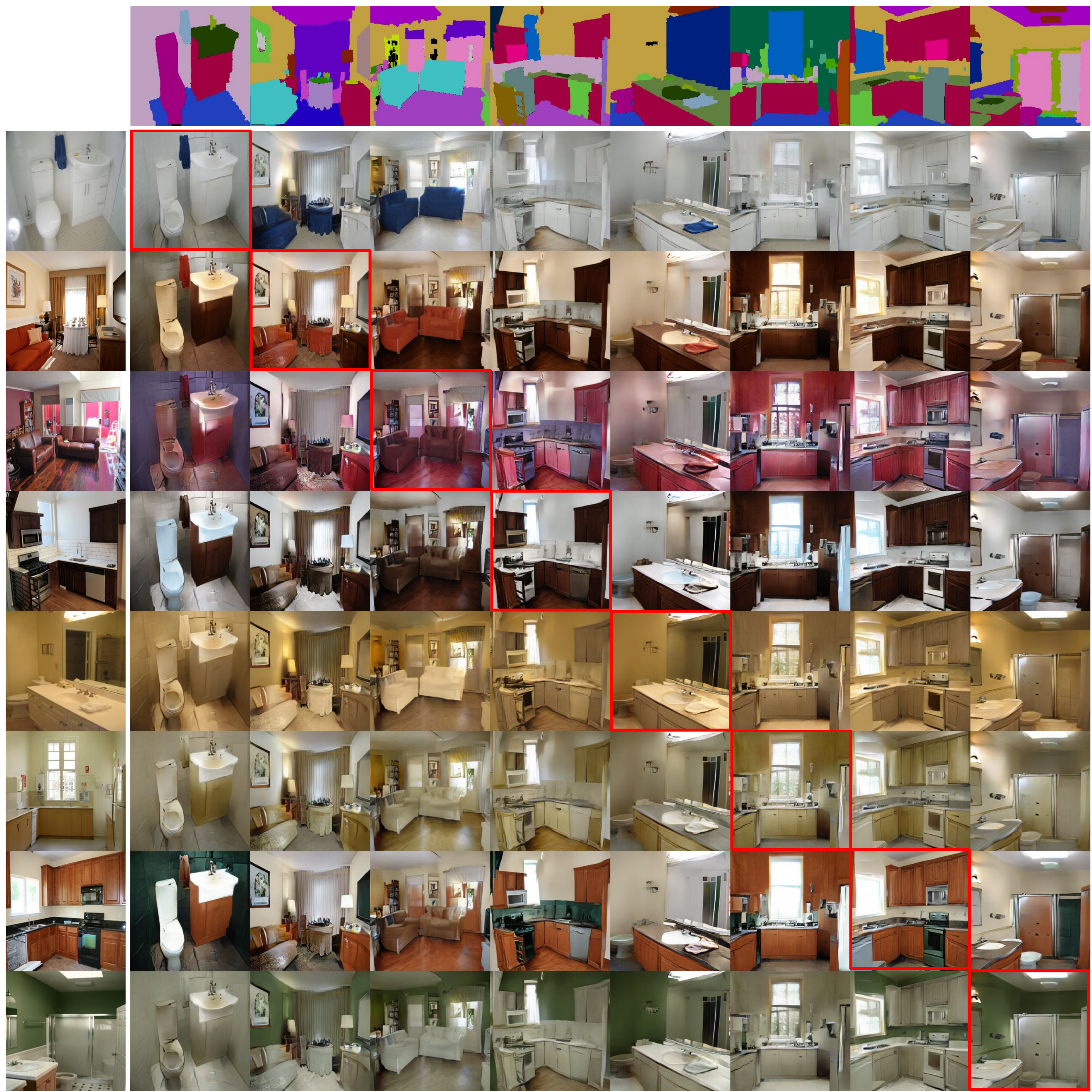}
	\caption{Style-structure swapping on 8 arbitrary indoor scenes at resolution $256\times 256$. 
    Our model can generalize across recognizably different scenes of different semantics, and synthesize images with reasonable and consistent styles. 
    Please zoom in for details.}
	\label{fig:swap_indoor}
\end{figure*}

We also train our model on an additional indoor scene dataset. Specifically, we collect 26713 diversified indoor scene images from the COCO-stuff dataset for training. As shown in Fig.~\ref{fig:swap_indoor}, our model is able to consistently transfer style across different indoor scenes, showing that our approach can effectively generalize to other datasets with complex structures. 

\vspace{-3px}
\section{More Scenes Swapping Results}
\label{sec:swap}
We show style swapping results on $12$ diversified scenes in Fig.~\ref{fig:swap1}. As shown in the figure, our model can transfer styles to very different scene semantics and generate style consistent outputs given exemplar images. Note that our model can infer the style of river (column 2) from other visual content of the exemplar images that does not contain water. Also, our model can precisely transfer styles for the foreground animals.

\begin{figure*}[]
	\centering
	\includegraphics[width=1.0\linewidth]{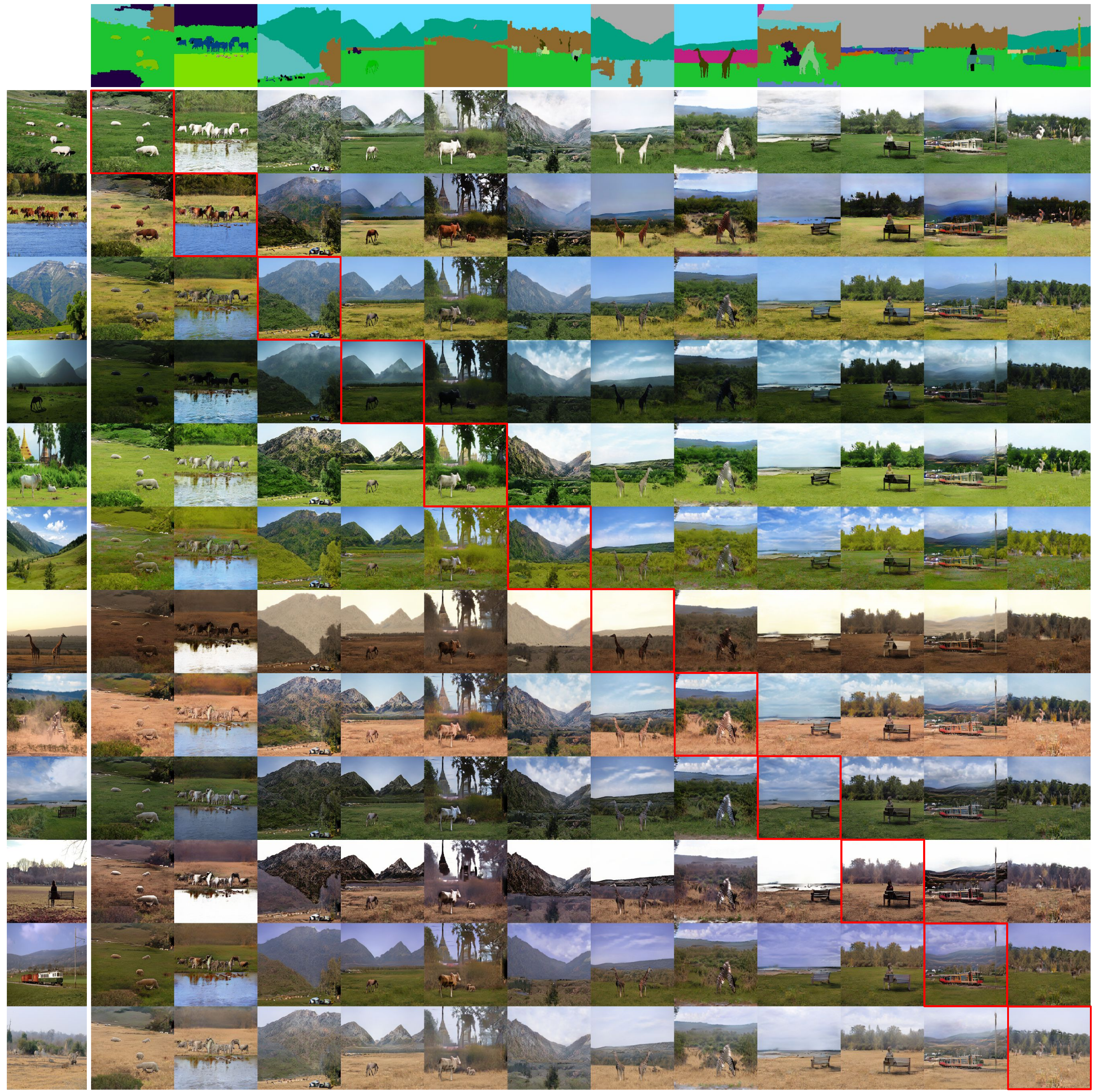}
	\caption{Style-structure swapping on 12 arbitrary scenes at resolution $256\times 256$. 
    Our model can generalize across recognizably different scenes of different semantics, and synthesize images with reasonable and consistent styles. 
    Our model can effectively transfer the color of foreground animals. 
    Also note that our model can implicitly infer the style of river (column 2) from exemplar images without relying on water-related visual contents.
    Please zoom in for details.}
	\label{fig:swap1}
\end{figure*}

\vspace{-3px}
\section{The Synthesis Module}
\label{sec:synthesis}
\begin{figure*}[h]
	\centering
	\includegraphics[width=1.0\linewidth]{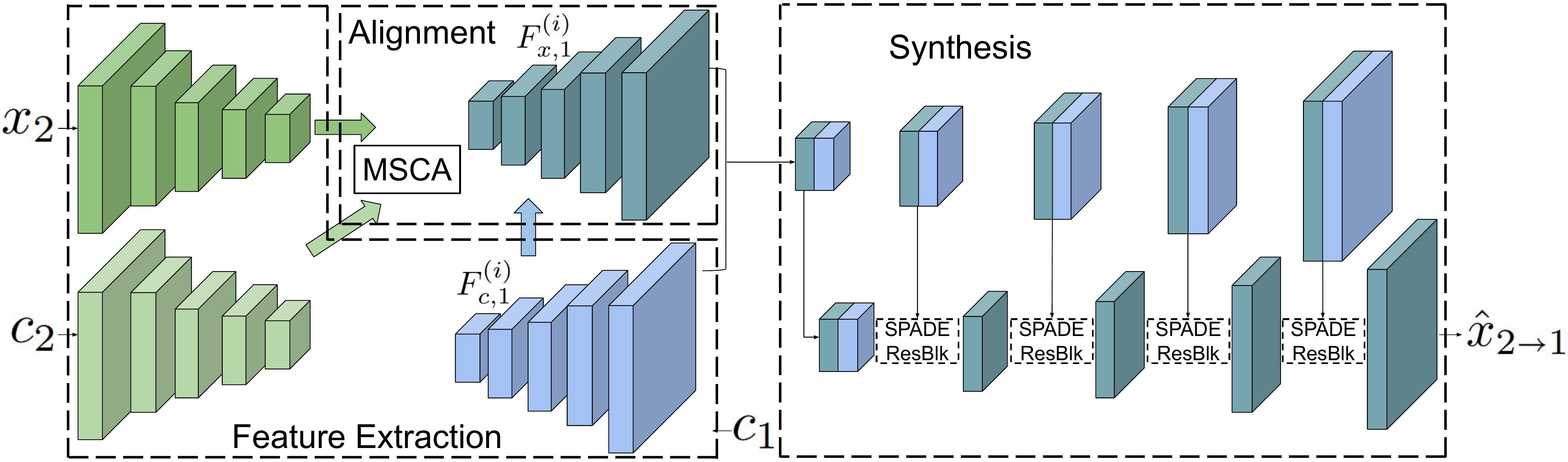}
	\caption{The details of the image synthesis module (the dash block on the right). The image synthesis module takes image features maps $F^{(i)}_{x,1}$ and segmentation features maps $F^{(i)}_{c,1}$ at all scale $i$ as inputs to output a new image $\hat{x}_{2\shortrightarrow 1}$. Multiple SPADE residue blocks~\cite{spade} with upsampling layers are used to upsample the spatial resolutions.
	}
	\label{fig:sup-synthesis}
\end{figure*}

\noindent As shown in Fig.~\ref{fig:sup-synthesis}, our image synthesis module (the dash block on the right) takes the image features map $F^{(i)}_{x,1}$ and segmentation features map $F^{(i)}_{c,1}$ as inputs to output a new image $\hat{x}_{2\shortrightarrow 1}$. Specifically, at each scale, a SPADE residue block~\cite{spade} with upsampling layer takes the concatenation of $F^{(i)}_{x,1}$ and $F^{(i)}_{c,1}$ as input to generate an upsampled feature map or image. 

\section{Patch Sampling Strategy and Training Details}
\label{sec:patch}
We employ the following patch sampling strategy during training. In the first $20$ epochs during training, we sample two $256\times 256$ patches from the $512\times512$ global images. Then we perform random collision avoidance in horizontal direction or vertical direction. To help our model generalize better across scenes that have larger variances, we employ only the patch-scale cross-reconstruction task at the first $20$ epochs.

During the next $20$ epochs of training, we randomly crop $384\times 384$ patches from the $512\times512$ global image, and resize patches to $256\times 256$ to generate patches for the patch-scale cross-reconstruction task. The enlarged patch scale can potentially helps our model to generalize at the global scale. 
In this stage, we employ global image to facilitate better global scale synthesis.
Random flipping is employed to augment data during training. 


\section{MSCA Pretraining}
\label{sec:pretraining}
\begin{figure*}[h]
	\centering
	\includegraphics[width=0.9\linewidth]{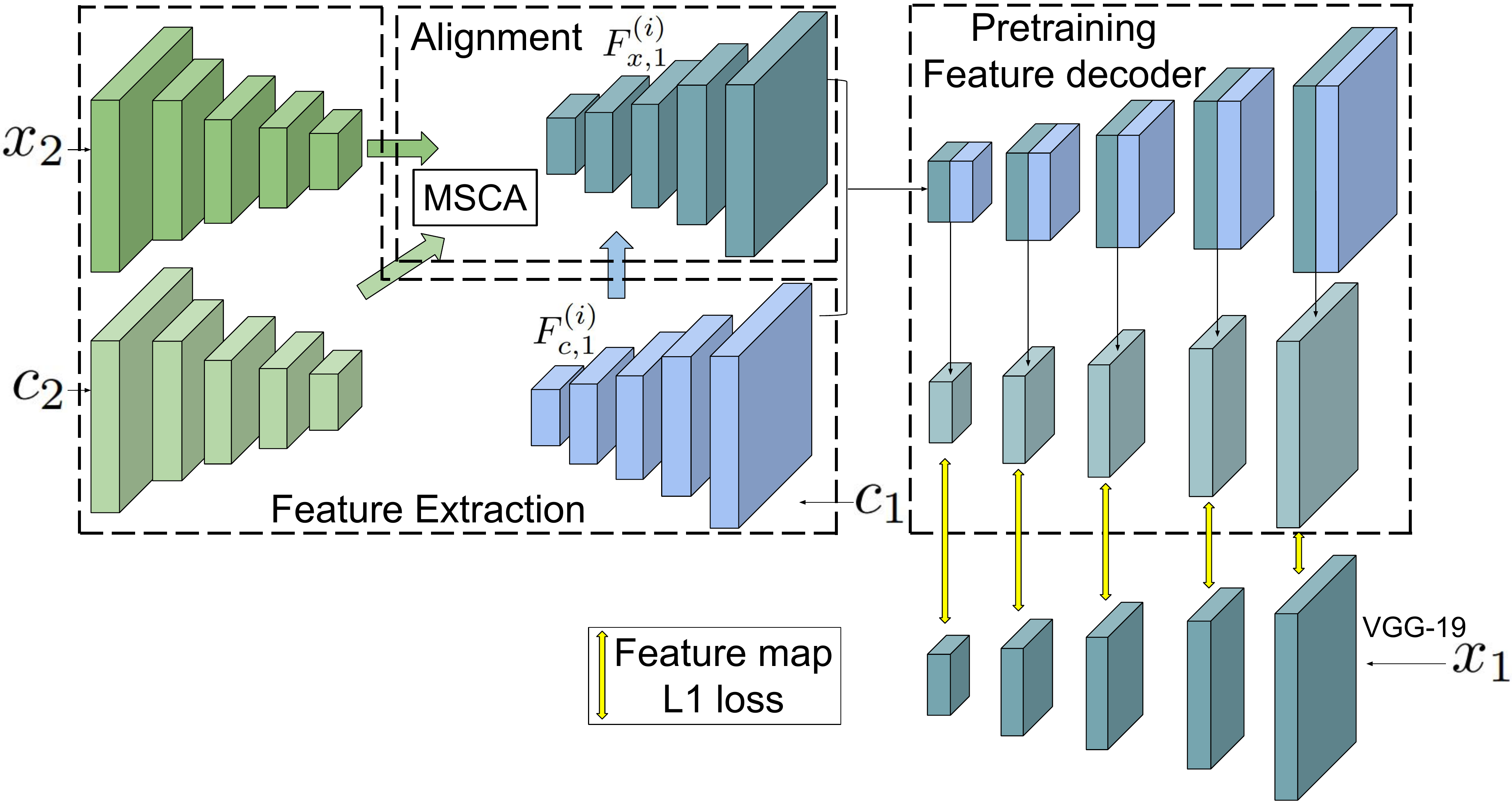}
	\caption{The details of the auxiliary feature decoder for feature extractor and MSCA pretraining (dash block on the right). At each scale $i$, the image features map $F^{(i)}_{x,1}$ and the segmentation features map $F^{(i)}_{c,1}$ are concatenated and feed to a $1\times 1$ convolution layer to predict the VGG-19 features map of $x_1$.
	}
	\label{fig:pretraining}
\end{figure*}

\noindent As shown in Fig.~\ref{fig:pretraining}, an auxiliary feature decoder (the dash block on the right) is used to pretrain the feature extractors and the MSCA modules. Specifically, at each scale, the concatenation of $F^{(i)}_{x,1}$ and $F^{(i)}_{c,1}$ at each scale is fed into a $1\times 1$ convolutional layer to reconstruct the ground-truth VGG feature of $x_1$ at the corresponding scale. We weighted sum the L1 losses between predictions and ground-truth at each scales, then apply backpropagation to update weights of the whole model. We pretrain the model for $20$ epochs. Because of the light-weight design of the feature decoder, the pretraining step only takes around $12$ hours, and around ~$2$\% of the total training time.
During the MSCA pretraining, we crop non-overlapping, $256\times 256$ patches from the $512\times512$ global images.

\end{document}